\documentclass[article]{llncs}

\usepackage[hidelinks,colorlinks=true,linkcolor=green,urlcolor=blue]{hyperref}
\usepackage{graphicx}
\usepackage{multirow}
\usepackage[linesnumbered]{algorithm2e}

\begin{document}

\title{Solving Logic Grid Puzzles with an Algorithm that Imitates Human Behavior\thanks{This publication has emanated from research conducted with the financial support of Science Foundation Ireland (SFI) under Grant Number SFI/12/RC/2289, co-funded under the European Regional Development Fund.}}

\author{Guillaume Escamocher \and Barry O'Sullivan}

\institute{Insight Centre for Data Analytics\\
University College Cork, Cork, Ireland\\
\email{\{guillaume.escamocher,barry.osullivan\}@insight-centre.org}}

\maketitle

\begin{abstract}
We present in this paper our solver for logic grid puzzles. The approach used by our algorithm mimics the way a human would try to solve the same problem. Every progress made during the solving process is accompanied by a detailed explanation of our program's reasoning. Since this reasoning is based on the same heuristics that a human would employ, the user can easily follow the given explanation.
\end{abstract}

\section{Introduction}

Most research work done in Computer Science aims to brings us closer to a machine that can solve any problem stated by any user. The most promising advancements towards this Holy Grail~\cite{holygrail} have come so far from Constraint Programming.

Nowadays Constraint Programming occurs in very different contexts, both in terms of the problems it is solving, and in terms of the people using it. Constraint Programming has been used by computer scientists to solve a multi-billion auction problem for the U.S. government involving thousands of television stations~\cite{radiospectrum}. It is also used, often unknowingly, by many people every day to solve the daily Sudoku puzzle from their favorite newspaper.

In this paper we focus on logic grid puzzles which, not unlike Sudoku, are casual problems. Following the instructions of the Challenge, we looked at the puzzles from the \href{http://logicgridpuzzles.com}{logicgridpuzzles.com} website. Throughout this paper, we will refer to individual puzzles from the website by their difficulty, their ID and their title in that order. So for example, the \href{http://logicgridpuzzles.com/puzzles/show_logic.php?ID=211}{Dragon Slayer} puzzle is Hard 211 ``Dragon Slayer''.

A logic puzzle is composed of $k$ \emph{categories}, of $k\times n$ \emph{elements} and of \emph{clues}. Categories are sets of elements, and all categories in the same puzzle contain the same number $n$ of elements. A \emph{solution} for a logic grid puzzle is a set of $n$ $k$-tuples, such that each tuple contains one element from each category, and such that no element occurs in distinct tuples. The clues are restrictions on the possible matchings. A valid logic grid puzzle has exactly one solution.

As an example, consider the following puzzle:
\begin{itemize}
\item 3 categories:
\begin{itemize}
\item ``First Name'', containing 3 elements (``Angela'', ``Donald'' and ``Leo'').
\item ``Country'', containing 3 elements (``Germany'', ``Ireland'' and ``United States'').
\item ``Year of Birth'', containing 3 elements (``1946'', ``1954'' and ``1979'').
\end{itemize}
\item 3 clues:
\begin{itemize}
\item The person from the ``United States'' was born in ``1946''.
\item ``Leo'' is younger than the person from ``Germany''.
\item ``Donald'' was born in ``1946'', or he is from ``Ireland''.
\end{itemize}
\end{itemize}
The only way to fulfill the clues is to match ``Angela'' with ``Germany'' and ``1954'', ``Donald'' with ``United States'' and ``1946'', and ``Leo'' with ``Ireland'' and ``1979''. This is the solution to the puzzle. Note that the solution can be found even though not all labels appear in the clues.

Our contribution is a constraint program that takes as input a logic grid puzzle and solves it while explaining the reasoning behind each variable assignment. Throughout the process, our algorithm behaves as a human trying to solve the same puzzle would.

The remainder of the paper is laid out as follows. Section~\ref{sec:implementation} describes how we implemented our approach. Its outline follows the four steps enumerated in the Challenge statement, with a fifth subsection presenting some other functions that we added to increase the customization possibilities of the explanation part. Section~\ref{sec:results} then details the characteristics of the puzzles we tested our program on, as well as the difficulties we encountered and how we dealt with them.

\section{Implementation}\label{sec:implementation}

Each of the following four subsections deals with the steps enumerated in the Challenge statement, in order. The last subsection of the current section presents features that we added to our program.

\subsection{Acquisition}\label{sec:input}

We are only partially addressing Step 1 of the Challenge. Instead of taking as input the website pages exactly as they are, our program asks the user to enter the clues as sets of constraints. For each constraint the user is adding, they are asked the clue to which the constraint belongs, which type of constraint is being added, and further information that is specific to each constraint type. We have implemented the following 14 constraints:

\begin{enumerate}
\item \emph{yes}: ``X'' is ``Y''. The user is asked to give two labels, corresponding to ``X'' and ``Y'' respectively. Example: Clue 2 in \href{http://logicgridpuzzles.com/puzzles/show_logic.php?ID=211}{Hard 211 ``Dragon Slayer''}.
\item \emph{no}: ``X'' is not ``Y''. The user is asked to give two labels, corresponding to ``X'' and ``Y'' respectively. Example: Clue 7 in \href{http://logicgridpuzzles.com/puzzles/show_logic.php?ID=107}{Hard 107 ``Bird Rescue 101''}.
\item \emph{or}: ``X'' is ``Y'' or ``Z''. The user is asked to give three labels, corresponding to ``X'', ``Y'' and ``Z'' respectively. Example: Clue 3 from Section 1's example.
\item \emph{xor}: ``X'' is either ``Y'' or ``Z''. The user is asked to give three labels, corresponding to ``X'', ``Y'' and ``Z'' respectively. Example: Clue 6 in \href{http://logicgridpuzzles.com/puzzles/show_logic.php?ID=107}{Hard 107 ``Bird Rescue 101''}.
\item \emph{alldiff}: ``X$_1$'', ``X$_2$'',\dots, ``X$_n$'' are all distinct. The user is first asked for the value of $n$, then asked to give $n$ labels, corresponding to the $n$ ``X$_i$''. Example: Clue 7 in \href{http://logicgridpuzzles.com/puzzles/show_logic.php?ID=71}{Hard 71 ``Home Sick''}.
\item \emph{twobytwo}: Out of ``X'' and ``Y'', one is ``W'' and the other is ``Z''. The user is asked to give four labels, corresponding to ``X'', ``Y'', ``W'' and ``Z'' respectively. Example: Clue 4 in \href{http://logicgridpuzzles.com/puzzles/show_logic.php?ID=74}{Hard 74 ``Class is in''}.
\item \emph{before}: ``X'' is before ``Y'' in the ``C'' category. The user is asked to give three labels, corresponding to ``X'', ``C'' and ``Y'' respectively. Example: Clue 6 in \href{http://logicgridpuzzles.com/puzzles/show_logic.php?ID=74}{Hard 74 ``Class is in''}.
\item \emph{after}: ``X'' is after ``Y'' in the ``C'' category. The user is asked to give three labels, corresponding to ``X'', ``C'' and ``Y'' respectively. Example: Clue 2 in \href{http://logicgridpuzzles.com/puzzles/show_logic.php?ID=93}{Hard 93 ``Pasta Night''}.
\item \emph{beforefixed}: ``X'' is exactly $n$ elements before ``Y'' in the ``C'' category. The user is first asked for the value of $n$, then asked to give three labels, corresponding to ``X'', ``C'' and ``Y'' respectively. Example: Clue 2 in \href{http://logicgridpuzzles.com/puzzles/show_logic.php?ID=107}{Hard 107 ``Bird Rescue 101''}.
\item \emph{afterfixed}: ``X'' is exactly  $n$ elements after ``Y'' in the ``C'' category. The user is first asked for the value of $n$, then asked to give three labels, corresponding to ``X'', ``C'' and ``Y'' respectively. Example: Clue 3 in \href{http://logicgridpuzzles.com/puzzles/show_logic.php?ID=74}{Hard 74 ``Class is in''}.
\item \emph{beforeatleast}: ``X'' is at least $n$ elements before ``Y'' in the ``C'' category. The user is first asked for the value of $n$, then asked to give three labels, corresponding to ``X'', ``C'' and ``Y'' respectively. Example: ``Laura paid at least 3 fewer dollars than the woman from Cork''.
\item \emph{afteratleast}: ``X'' is at least $n$ elements after ``Y'' in the ``C'' category. The user is first asked for the value of $n$, then asked to give three labels, corresponding to ``X'', ``C'' and ``Y'' respectively. Example: ``The movie that Emily rented came out at least 10 years after \emph{Fight Club}''.
\item \emph{distance}: ``X'' is exactly $n$ elements from ``Y'' in the ``C'' category. The user is first asked for the value of $n$, then asked to give three labels, corresponding to ``X'', ``C'' and ``Y'' respectively. Example: ``The Norwegian lives next to the blue house'' in the \href{https://en.wikipedia.org/wiki/Zebra_Puzzle}{Zebra} puzzle.
\item \emph{disjunction}: ``X$_1$'' is ``Y$_1$'' or ``X$_2$'' is not ``Y$_2$'' or \dots or ``X$_n$'' is ``Y$_n$''. The user is first asked for the number of disjuncts, then for the polarity of each of them (``is'' is positive, ``is not'' is negative) then finally they are asked to give $2n$ labels, corresponding to ``X$_1$'', ``Y$_1$'',\dots,``Y$_n$'' in order. Example: ``If the 23-year-old person is wearing a blue shirt, then Bill did not order a burger''.
\end{enumerate}

For each constraint, the order in which the labels are requested mirrors the order in which they generally appear in clues of this type. In particular, this is the reason why the category label must be entered between the element labels in Constraints 7-13.

One might think there is too much redundancy with that many types of constraints. Some types are particular cases of others (for example 7 is the same as 11 with $n$ set to 1), some are dual of others (9 is the same as 10 with the order of the labels reversed) and some can be written as a conjunction of others (all types can be written as a conjunction of type 14 constraints). We chose to implement that many constraints in order to be as close as possible to the original formulation of the clues. The only constraint type that deviates from this intent is 14, which was added to cover as many possible clues as possible.

To illustrate this part of our implementation, this is what the constraints from Section 1's example look like after input is taken:
\begin{itemize}
\item \{1,yes,``United States'',``1946''\}
\item \{2,after,``Leo'',``Year of Birth'',``Germany''\}
\item \{3,or,``Donald'',``1946'',``Ireland''\}
\end{itemize}

Each constraint contains as part of its description the clue it belongs to. Indeed, clues can contain several constraints (see for example \href{http://logicgridpuzzles.com/puzzles/show_logic.php?ID=8}{Easy 8 ``A Michigan Adventure''}), and by referring to the clue instead of the constraint the explanation is more clear for the user.

\subsection{Model}

The ``grid'' part in the name of this puzzle genre refers to the representation most often used to solve them. Usually for a logic grid puzzle with $k$ categories, the elements from Categories 2 to $k$ label the columns, while the elements from Category 1 and Categories $k$ to 3 label the rows. The exact positioning of the categories does not actually matter, as long as each pair of elements from different categories is represented by exactly one cell. Figure~\ref{fig:examplegrid} shows the grid corresponding to the running example.

\begin{figure}
\centering
\def\arraystretch{1.5}
\begin{tabular}{|c|l||c|c|c||c|c|c|}
\cline{3-8}
\multicolumn{2}{c|}{} & \multicolumn{3}{c||}{Country} & \multicolumn{3}{c|}{Year of Birth}\\
\cline{3-8}
\multicolumn{2}{c|}{} & \rotatebox{90}{Germany} & \rotatebox{90}{Ireland} & \rotatebox{90}{United States} & \rotatebox{90}{1946} & \rotatebox{90}{1954} & \rotatebox{90}{1979}\\
\hline
\hline
\multirow{3}{*}{\rotatebox{90}{\scalebox{.9}{First Name}}} & Angela & \hspace{.2in} & \hspace{.2in} & \hspace{.2in} & \hspace{.2in} & \hspace{.2in} & \hspace{.2in}\\
\cline{2-8}
& Donald & & & & & &\\
\cline{2-8}
& Leo & & & & & &\\
\hline
\hline
\multirow{3}{*}{\rotatebox{90}{\scalebox{.9}{Year of Birth}}} & 1946 & & & &\multicolumn{3}{c}{}\\
\cline{2-5}
& 1954 & & & &\multicolumn{3}{c}{}\\
\cline{2-5}
& 1979 & & & &\multicolumn{3}{c}{}\\
\cline{1-5}
\end{tabular}
\caption{The running example represented as a grid.}
\label{fig:examplegrid}
\end{figure}

Solving a puzzle consists in filling each cell of its grid by either \emph{yes} if the label of the row is matched with the label of the column in the solution, or by \emph{no} otherwise. In our model, we represent the grid by an array of integers. Each cell is initially set to 0. Once a pair of elements is determined to be matched in the solution, the corresponding cell in the array is set to 1. Similarly, if two elements from different categories are determined to not be matched together in the solution, then the corresponding cell in the array is set to -1.

\subsection{Solving}

Our main goal when addressing Step 3 of the Challenge was to create a solver that behaves exactly as a human would. Therefore our solver uses the same inference rules, in the same order, as the user would if trying to solve the puzzle.

At each step of the process, our algorithm picks a rule from a set of inference rules and applies it to fill one cell. The rules can be divided into two kinds: consistency rules that only use information from the current state of the grid, and clues rules that also use information from the constraints forming the clues. Consistency rules can be further divided between basic consistency rules that are easy to use for humans, and advanced consistency rules that are more complicated and will be avoided by humans unless necessary. Some of our consistency rules correspond to previously discovered inference rules~\cite{oldpuzzles}.

Basic consistency rules use the meta-information of logic grid puzzles: for any two distinct categories $C$ and $D$, there is a bijection between the elements of $C$ and the elements of $D$. So if in the running example we know that the cell (``1946'',``United States'') is filled with \emph{yes}, then the cells (``1954'',``United States'') and (``1979'',``United States'') must be filled with \emph{no}. Also, if the cells (``Leo'',``1946'') and (``Leo'',``1954'') have already been filled with \emph{no}, then (``Leo'', ``1979''), the last cell for ``Leo'' in the ``Year of Birth'' category, must be filled with \emph{yes}.

Our most frequently used advanced consistency rule relies on the transitivity inherent from the grid format. So if in the running example both (``Donald'',``1946'') and (``1946'',``United States'') are filled with \emph{yes}, then (``Donald'',``United States'') must also be filled with \emph{yes}. Another one of our advanced consistency rules bears similarities with path consistency. If we have two elements from different categories labelled ``e'' and ``f'', and a third category $C$ such that for each element in $C$ labelled ``g'' either the cell (``e'',``g'') or the cell (``f'',``g'') is filled with \emph{no}, then it means that no element of $C$ can be matched with both ``e'' and ``f'', so ``e'' and ``f'' cannot be matched together in the solution, and therefore the cell (``e'',``f'') can be filled with \emph{no}. 

Each rule that uses information from the clues is associated with one of our fourteen constraints. Some of the constraints are only associated with one trivial rule. For example the rule associated with Constraint 1 (``X'' is ``Y'') directly fills the cell (``X'',``Y'') with \emph{yes}. Other constraints are associated with several, more elaborate rules that not only use the information from the clue the constraint is part of, but also take into account the current state of the grid. For example one of the rules for Constraint 3 (``X'' is ``Y'' or ``Z'') fills the cell (``X'',``Z'') with \emph{yes} if the cell (``X'',``Y'') contains \emph{no}, and another rule for the same constraint fills the cell (``X'',``W'') with \emph{no} for every label ``W'' such that both (``W'',``Y'') and (``W'',``Z'') contain \emph{no}.

Humans will naturally read the clues in order and fill what they can from that information. Then they will complete rows and columns in categories where either one cell is filled with \emph{yes} or all cells but one are filled with \emph{no}. Once they have exhausted all easy ways to make progress, and only then, they will use the more complicated reasoning found in our advanced consistency rules. When deciding which rule to try next, our solver reflects that method of thinking.

\begin{algorithm}
\caption{\label{alg:solver}Logic grid puzzle solver.}
\KwData{A logic grid puzzle.}
\KwResult{The solution to the puzzle in the form of a filled grid.}
$progress\leftarrow true$\;
\While{progress}
	{$progress\leftarrow false$\;
	\For{$i\leftarrow 1$ \KwTo $|Cons|$}
		{$rule\leftarrow nextrule(Cons)$\;
		\If{$canapply(rule)$}
			{$apply(rule)$\;
			$progress\leftarrow true$\;}}
	$easyprogress\leftarrow true$\;
	\While{$easyprogress$}
		{$easyprogress\leftarrow false$\;
		\For{$i\leftarrow 1$ \KwTo $|BCR|$}
			{$rule\leftarrow nextrule(BCR)$\;
			\If{$canapply(rule)$}
				{$apply(rule)$\;
				$progress\leftarrow true$\;
				$easyprogress\leftarrow true$\;}}}
	\If{$progress==false$}
		{$difficultprogress\leftarrow false$\;
		\While{$difficultprogress==false$}
			{$rule\leftarrow nextrule(ACR)$\;
			\If{$canapply(rule)$}
				{$apply(rule)$\;
				$difficultprogress\leftarrow true$\;}}}}
\end{algorithm}

The priority order between the different kinds of inference rules is illustrated in Algorithm~\ref{alg:solver}. $Cons$ is the set of rules using information from the clues, $BCR$ is the set of basic consistency rules and $ACR$ is the set of advanced consistency rules. As can be seen in Lines 4-10, the algorithm tries the constraints in order, without immediately trying earlier constraints again when progress is made. On the other hand, basic consistency rules are used over and over because of their extreme simplicity, which makes them very attractive for a human user. Finally, advanced consistency rules are only used as a last resort, when no further progress is possible from any other rule. 

\subsection{Explanation}

Every time a cell is filled, our program outputs the reasoning that led it to determine the correct value for that particular cell. So if called on the running example, our program writes the following when using the information from the clues to start solving the puzzle:

\begin{itemize}
\item ``United States'' is ``1946'' (Clue 1).
\item ``Leo'' is after ``Germany'' in the ``Year of Birth'' category (Clue 2), so ``Leo'' is not ``Germany''.
\item ``Leo'' is after ``Germany'' in the ``Year of Birth'' category (Clue 2), so ``Leo'' is not the first element in that category, so ``Leo'' is not ``1946''.
\item ``Leo'' is after ``Germany'' in the ``Year of Birth'' category (Clue 2), so ``Germany'' is not the last element in that category, so ``Germany'' is not ``1979''.
\end{itemize}

Note that Clue 3 cannot be exploited yet. The state of the grid at this point is pictured in Figure~\ref{fig:grid1}, with ``Y'' representing \emph{yes} and a dot representing \emph{no}.

\begin{figure}
\centering
\def\arraystretch{1.5}
\begin{tabular}{|c|l||c|c|c||c|c|c|}
\cline{3-8}
\multicolumn{2}{c|}{} & \multicolumn{3}{c||}{Country} & \multicolumn{3}{c|}{Year of Birth}\\
\cline{3-8}
\multicolumn{2}{c|}{} & \rotatebox{90}{Germany} & \rotatebox{90}{Ireland} & \rotatebox{90}{United States} & \rotatebox{90}{1946} & \rotatebox{90}{1954} & \rotatebox{90}{1979}\\
\hline
\hline
\multirow{3}{*}{\rotatebox{90}{\scalebox{.9}{First Name}}} & Angela & \hspace{.2in} & \hspace{.2in} & \hspace{.2in} & \hspace{.2in} & \hspace{.2in} & \hspace{.2in}\\
\cline{2-8}
& Donald & & & & & &\\
\cline{2-8}
& Leo & \scalebox{.8}{$\bullet$} & & & \scalebox{.8}{$\bullet$} & &\\
\hline
\hline
\multirow{3}{*}{\rotatebox{90}{\scalebox{.9}{Year of Birth}}} & 1946 & & & Y &\multicolumn{3}{c}{}\\
\cline{2-5}
& 1954 & & & &\multicolumn{3}{c}{}\\
\cline{2-5}
& 1979 & \scalebox{.8}{$\bullet$} & & &\multicolumn{3}{c}{}\\
\cline{1-5}
\end{tabular}
\caption{Using the clues to start the solving process.}
\label{fig:grid1}
\end{figure}

The solver now has enough information to complete the bottom left block:

\begin{itemize}
\item 7 cells can be filled from basic consistency.
\end{itemize}

To avoid cluttering the explanation, the solver groups together consecutive lines of basic consistency application. This is entirely optional, and can be turned off. Now that basic consistency has filled more cells, the solver tries using the clues again to check whether some new information can be inferred:

\begin{itemize}
\item ``Germany'' is not one of the first 1 element in the ``Year of Birth'' category, and ``Leo'' is after ``Germany'' in that category (Clue 2), so ``Leo'' is not one of the first 2 elements in the ``Year of Birth'' category, so ``Leo'' is not ``1954''.
\item ``Donald'' is ``1946'' or ``Ireland'' (Clue 3), and ``Germany'' is neither ``1946'' nor ``Ireland'', so ``Donald'' is not ``Germany''.
\item ``Donald'' is ``1946'' or ``Ireland'' (Clue 3), and ``1954'' is neither ``1946'' nor ``Ireland'', so ``Donald'' is not ``1954''.
\item 9 cells can be filled from basic consistency.
\end{itemize}

\begin{figure}
\centering
\def\arraystretch{1.5}
\begin{tabular}{|c|l||c|c|c||c|c|c|}
\cline{3-8}
\multicolumn{2}{c|}{} & \multicolumn{3}{c||}{Country} & \multicolumn{3}{c|}{Year of Birth}\\
\cline{3-8}
\multicolumn{2}{c|}{} & \rotatebox{90}{Germany} & \rotatebox{90}{Ireland} & \rotatebox{90}{United States} & \rotatebox{90}{1946} & \rotatebox{90}{1954} & \rotatebox{90}{1979}\\
\hline
\hline
\multirow{3}{*}{\rotatebox{90}{\scalebox{.9}{First Name}}} & Angela & Y & \hspace{.075in}\scalebox{.8}{$\bullet$}\hspace{.075in} & \hspace{.075in}\scalebox{.8}{$\bullet$}\hspace{.075in} & \hspace{.075in}\scalebox{.8}{$\bullet$}\hspace{.075in} & Y & \scalebox{.8}{$\bullet$}\\
\cline{2-8}
& Donald & \hspace{.075in}\scalebox{.8}{$\bullet$}\hspace{.075in} & & & Y & \hspace{.075in}\scalebox{.8}{$\bullet$}\hspace{.075in} & \scalebox{.8}{$\bullet$}\\
\cline{2-8}
& Leo & \scalebox{.8}{$\bullet$} & & & \scalebox{.8}{$\bullet$} & \scalebox{.8}{$\bullet$} & Y\\
\hline
\hline
\multirow{3}{*}{\rotatebox{90}{\scalebox{.9}{Year of Birth}}} & 1946 & \scalebox{.8}{$\bullet$} & \scalebox{.8}{$\bullet$} & Y &\multicolumn{3}{c}{}\\
\cline{2-5}
& 1954 & Y & \scalebox{.8}{$\bullet$} & \scalebox{.8}{$\bullet$} &\multicolumn{3}{c}{}\\
\cline{2-5}
& 1979 & \scalebox{.8}{$\bullet$} & Y & \scalebox{.8}{$\bullet$} &\multicolumn{3}{c}{}\\
\cline{1-5}
\end{tabular}
\caption{The solution is almost complete.}
\label{fig:grid2}
\end{figure}

Figure~\ref{fig:grid2} shows that at this point the grid is almost completely filled. However the clues are now all fully satisfied, and all possibilities of progress from basic consistency rules have been exhausted. Therefore the algorithm must use an advanced transitivity rule to take the next step towards the solution:

\begin{itemize}
\item ``Donald'' is ``1946'' and ``1946'' is ``United States'', so ``Donald'' is ``United States''.
\item 3 cells can be filled from basic consistency.
\end{itemize}

By default our program outputs the full explanation at once, however the option to only write one line at a time has also been implemented. This can be useful if the user wants to solve the puzzle by themselves, but is stuck at one stage and desires a hint.

\subsection{Additional Features}

We have enhanced our program with some additional functions to offer a more customizable explanation to the user. One of these lets the user know every time they can discard a clue. While technically this corresponds to the last time that clue is used, which can be easily determined by checking the last line that the clue appears in the explanation, we instead look for the moment when already filled cells of the grid explicitly indicate that the clue has been fully satisfied. This choice keeps with the general intent of our work, which is to build a solver that reasons as a human would.

In most logic grid puzzles, the aim is to fill the entire grid. It is however conceivable that a user would be interested in only knowing the value of one single cell. In fact, the famous \href{https://en.wikipedia.org/wiki/Zebra_Puzzle}{Zebra} puzzle does not ask for the matching of all elements, but only for who in the ``Nationality'' category is paired with the eponymous ``Zebra''. If requested to do so, our program can restrict the explanation to only the part which is relevant to the determination of a particular cell value.

Finally, our program can convert logic grid puzzles into Conjunctive Normal Form (CNF). The resulting CNF files can then be used by any SAT solver or model counter. We primarily implemented this feature as a debugging tool, to check the validity of new puzzles.

\section{Results}\label{sec:results}

\subsection{Scope}

Our tests encompass the puzzles that are presented as either easy or hard on the website, as well as the \href{https://en.wikipedia.org/wiki/Zebra_Puzzle}{Zebra} puzzle, probably the first and most well-known puzzle of this type.

We removed 8 puzzles from our consideration. One of them (Hard 72 ``Expensive Coffee'') has a clue that refers to a label not present in the corresponding category. Editing this label to one of the five elements in that particular category led to no solution in three cases, and to a unique but rejected solution in the other two cases. The clues in the seven other puzzles that we removed (Easy 23 ``Football fanatics'', Easy 57 ``Special Delivery'', Easy 75 ``Three Friends'', Easy 108 ``Movie Buffs Associated Week of Films - Helen Mirren'', Easy 133 ``Easter Eggs'', Hard 73 ``For sale... sold!'', Hard 127 ``Secret Santa'') were not enough to reduce the number of solutions to 1. This can be easily checked manually for the easy puzzles, while for the hard ones we used our CNF conversion feature and applied a SAT model counter on the resulting file for confirmation.

After removing these defective puzzles, we were left with 69 puzzles: 55 of easy difficulty, 13 of hard difficulty, and the Zebra puzzle. All but 3 of them have 4 categories, 2 of them have 5 categories and the last one (the Zebra puzzle) has 6 categories. The number of elements in each category is 3 for 50 puzzles, 4 for 6 puzzles, and 5 for the other 13 puzzles.

For six puzzles (\href{http://logicgridpuzzles.com/puzzles/show_logic.php?ID=22}{Easy 22 ``Baggage Mishaps''}, \href{http://logicgridpuzzles.com/puzzles/show_logic.php?ID=60}{Easy 60 ``Holiday Decision''}, \href{http://logicgridpuzzles.com/puzzles/show_logic.php?ID=64}{Easy 64 ``Robbery at Millionaire's Mansion''}, \href{http://logicgridpuzzles.com/puzzles/show_logic.php?ID=70}{Easy 70 ``The racehorses''}, \href{http://logicgridpuzzles.com/puzzles/show_logic.php?ID=76}{Easy 76 ``Three little boys''}, \href{http://logicgridpuzzles.com/puzzles/show_logic.php?ID=83}{Easy 83 ``The Enchanted Forest''}), the information contained in the opening statement was needed to have no more than one solution. To address this, we simply treated the statement as an additional clue (Clue 0).

\subsection{Successes and Challenges}

Out of the 69 puzzles considered, 67 are straightforward to model by our set of fourteen constraints. The clues in one of the other two puzzles, \href{http://logicgridpuzzles.com/puzzles/show_logic.php?ID=119}{Hard 119 ``A New Personal Computer''}, contain the label ``Andrew'' which does not fit in any of the existing categories, despite being directly part of the puzzle main objective (``Which computer has been chosen by Andrew?''). We resolved this issue by adding a new ``Andrew'' category containing the labels ``Andrew'', ``NotAndrew1'', ``NotAndrew2'',  ``NotAndrew3'' and  ``NotAndrew4''. This allowed us to fully model all clues in that puzzle, and subsequently to solve it. We also added ordering constraints (by using Constraint 7) on the four ``NotAndrew'' labels, in order to keep the uniqueness of the solution. This has no effect on the answer to the original puzzle question.

The clues in the last puzzle, \href{http://logicgridpuzzles.com/puzzles/show_logic.php?ID=65}{Easy 65 ``Sporting Excellence''}, contain cross-referencing meta-information. Our representation of a clue as a conjunction of constraints was not able to model them individually.

Our algorithm managed to find the unique solution, accompanied by a human-readable explanation, for all 68 puzzles we could model in the input part of the Challenge. This gives us a success rate of $98.6\%$. Our only failure was on an Easy puzzle with only four categories and three elements in each category, which proves that the issue has nothing to do with scale.

On a Dell laptop with an Ubuntu 18.04 operating system and an Intel i7-5600U processor, solving all 68 puzzles takes a combined time of 91 milliseconds with file logging, 71 milliseconds without. This shows that our method is extremely computationally efficient.

\end{document}